\documentclass[11pt]{article}
\usepackage{acl}
\usepackage{graphicx}
\usepackage{booktabs}
\usepackage{amsmath}
\usepackage{amssymb}
\usepackage{hyperref}
\usepackage[T1]{fontenc}
\usepackage[utf8]{inputenc}
\usepackage{lmodern}

\title{More Rounds, More Noise: Why Multi-Turn Review Fails\\to Improve Cross-Context Verification}

\author{Song Tae-Eun \\
  Daejeon Jungang Cheonggua Co., Ltd. \\
  \texttt{higheun@gmail.com}}

\begin{document}
\maketitle

\begin{abstract}
Cross-Context Review (CCR) improves LLM verification by separating production and review into independent sessions.
A natural extension is multi-turn review: letting the reviewer ask follow-up questions, receive author responses, and review again.
We call this Dynamic Cross-Context Review (D-CCR).
In a controlled experiment with 30 artifacts and 150 injected errors, we tested four D-CCR variants against the single-pass CCR baseline.
Single-pass CCR (F1 = 0.376) significantly outperformed all multi-turn variants, including D-CCR-2b with question-and-answer exchange (F1 = 0.303, $p < 0.001$, $d = -0.59$).
Multi-turn review increased recall (+0.08) but generated 62\% more false positives (8.5 vs.\ 5.2), collapsing precision from 0.30 to 0.20.
Two mechanisms drive this degradation: (1) \emph{false positive pressure}---reviewers in later rounds fabricate findings when the artifact's real errors have been exhausted, and (2) \emph{Review Target Drift}---reviewers provided with prior Q\&A exchanges shift from reviewing the artifact to critiquing the conversation itself.
Independent re-review without prior context (D-CCR-2c) performed worst (F1 = 0.263), confirming that mere repetition degrades rather than helps.
The degradation stems from false positive pressure in additional rounds, not from information amount---within multi-turn conditions, more information actually helps (D-CCR-2b $>$ D-CCR-2a).
The problem is not what the reviewer sees, but that reviewing again invites noise.
\end{abstract}

\section{Introduction}

Cross-Context Review \citep[CCR;][]{song2026ccr} demonstrated that LLM verification improves when production and review happen in separate sessions.
By removing the production context, the reviewer avoids anchoring, sycophancy, and context degradation---biases that plague same-session self-review.
CCR achieved an F1 of 28.6\%, significantly outperforming same-session review (24.6\%, $p = 0.008$) and repeated same-session review (21.7\%, $p < 0.001$).

Can we do better with multiple rounds? Human review processes routinely involve follow-up questions---a code reviewer asks why a particular approach was chosen; a peer reviewer requests clarification on methodology.
These exchanges often surface issues that a single read misses.
\citet{kim2025interviewer} showed that multi-turn interviewer-style LLM evaluation reduces bias and improves robustness compared to single-pass assessment, though without context separation.

We tested whether combining CCR's context separation with multi-turn interaction yields further gains.
In Dynamic Cross-Context Review (D-CCR), the reviewer conducts an initial review, generates follow-up questions, receives the author's answers in a separate session preserving context separation, and reviews again with this additional information.

The results are negative.
Every multi-turn variant produced lower F1 than the single-pass baseline, with three of four comparisons highly significant ($p < 0.001$).
Additional rounds generate false positives faster than they discover true errors.

This paper makes four contributions:

\begin{enumerate}
\item We demonstrate empirically that the optimal number of CCR review rounds is one.
Multi-turn review significantly degrades F1 ($-0.073$, $p < 0.001$, $d = -0.59$) despite modestly improving recall (+0.08).

\item We identify \emph{false positive pressure}---a systematic tendency for reviewers to fabricate findings in later rounds when the artifact's real errors have been largely exhausted in round one.

\item We document \emph{Review Target Drift}---a failure mode where reviewers provided with Q\&A exchanges shift their attention from the artifact to the conversation, generating findings about the author's answers rather than the artifact itself.

\item We establish a \emph{round optimization principle}: within multi-turn conditions, more information helps (D-CCR-2b $>$ D-CCR-2a), but any second round degrades performance relative to a single pass.
The problem is not what the reviewer sees, but that reviewing again invites noise.
\end{enumerate}

\section{Related Work}

\subsection{Multi-Round LLM Verification}

Several lines of work have explored iterative LLM verification.
\citet{guo2025temporal} showed that temporal consistency checking improves error identification in mathematical reasoning, but within a single context.
\citet{chen2025magicore} proposed MAgICoRe, a multi-agent iterative refinement framework where Solver, Reviewer, and Refiner agents collaborate in repeated loops, achieving gains over self-consistency and self-refinement baselines.
The critical difference: MAgICoRe's agents share context, while D-CCR maintains strict context separation.

\citet{shi2025ssr} developed Socratic Self-Refine (SSR), decomposing reasoning into verifiable sub-question/sub-answer pairs and iteratively re-solving unstable steps.
The Socratic probing approach parallels D-CCR's follow-up question mechanism, but SSR operates within a single session.
\citet{du2025thinkthrice} showed that progressive thought refinement across multiple passes improves reasoning, though via fine-tuning rather than inference-time intervention.

These works establish that multi-round verification \emph{can} help within shared or continuous contexts. We show that when context separation is maintained, the dynamics change fundamentally.

\subsection{Self-Correction Limitations}

The self-correction literature explains why additional rounds might fail.
\citet{kumar2024score} demonstrated that prompting-based self-correction is fundamentally limited---reinforcement learning is needed for genuine improvement.
Giving a CCR reviewer another chance, even with additional information, does not overcome these limitations.

\citet{huang2024llms} and \citet{kamoi2024when} established that LLMs cannot reliably correct their own reasoning without external feedback.
\citet{zhang2025dark} found that self-correction can flip correct answers to incorrect ones.
\citet{tsui2025selfcorrection} identified the self-correction blind spot: LLMs fail to correct errors in their own outputs while succeeding on identical errors presented as external input.
CCR exploits this asymmetry; our results suggest that additional rounds erode the advantage.

\subsection{Multi-Agent Debate}

Multi-agent debate (MAD) research illuminates why more interaction does not always help.
\citet{zhang2025mad} evaluated five MAD methods across nine benchmarks and found that MAD does not reliably outperform single-agent baselines.
The only consistent improvement factor was \emph{model heterogeneity}---using different models, not different prompts.
D-CCR uses a single model, which may explain why additional rounds fail to introduce genuine diversity.

\citet{wu2025debate} identified \emph{majority pressure} as a key failure mode: when agents interact, the majority view suppresses independent correction.
This parallels our Review Target Drift---D-CCR reviewers who see prior Q\&A exchanges shift toward evaluating the discussion rather than independently examining the artifact.

\citet{smit2024mad} confirmed that MAD does not reliably beat simpler methods like self-consistency or ensembling.
\citet{gou2024critic} showed that external tools (calculators, code execution) enable genuine self-correction in ways that purely verbal feedback cannot---suggesting that D-CCR's verbal Q\&A exchange is insufficient as ``external feedback.''

\subsection{CCR Foundation}

\citet{song2026ccr} established that context separation is the operative mechanism in CCR: reviewing in a fresh session (F1 = 28.6\%) outperforms same-session review (24.6\%, $p = 0.008$), and repeated same-session review provides no benefit (SR2 $\approx$ SR, $p = 0.11$).
We extend this by testing whether the \emph{number} of separated review rounds matters.
That one round is optimal strengthens the original claim: it is the separation itself, not the amount of review, that drives improvement.

\section{Method: Dynamic Cross-Context Review}

\subsection{D-CCR Protocol}

D-CCR extends CCR by adding structured interaction between reviewer and author across separated sessions:

{\small
\begin{verbatim}
CCR (single-pass):
  A -> Artifact -> B -> Report

D-CCR (multi-turn):
  A -> Artifact -> B1 -> Review + Questions
                              |
                   A' -> Answers
                              |
                   B2 -> Final Report
\end{verbatim}
}

Each arrow represents a separate session with no shared context.
B$_1$ and B$_2$ are independent reviewer sessions.
A$'$ is a new author session that sees only the artifact and the questions---not the original production context.

\subsection{Experimental Conditions}

We tested four conditions, varying what information the reviewer receives in round two:

\begin{table}[h]
\centering
\resizebox{\columnwidth}{!}{%
\begin{tabular}{llll}
\toprule
\textbf{Condition} & \textbf{Round 1} & \textbf{Round 2} & \textbf{Purpose} \\
\midrule
CCR-1 & Artifact only & --- & Baseline \\
D-CCR-2a & Artifact only & Artifact + questions & Question-only \\
D-CCR-2b & Artifact only & Artifact + Q\&A & Full D-CCR \\
D-CCR-2c & Artifact only & Artifact only (fresh) & Independence \\
\bottomrule
\end{tabular}%
}
\end{table}

D-CCR-2a isolates the effect of the reviewer seeing their own prior questions.
D-CCR-2b is the full D-CCR protocol with author answers.
D-CCR-2c controls for the ``two looks'' effect by providing a completely fresh second review with no information from round one.

\subsection{Research Questions}

\textbf{RQ1.} Does multi-turn D-CCR outperform single-pass CCR? (H1: D-CCR-2b $>$ CCR-1)

\textbf{RQ2.} Does the author's answer help or anchor the reviewer? (H3: D-CCR-2b vs.\ D-CCR-2a)\footnote{H2 (D-CCR-3b vs.\ D-CCR-2b) was dropped after pilot---see Appendix~\ref{app:pilot}.}

\textbf{RQ3.} Is continuity better than independence in multi-round review? (H4: D-CCR-2b vs.\ D-CCR-2c)

\textbf{RQ4.} Does independent repetition improve over single-pass? (H5: D-CCR-2c vs.\ CCR-1)

\section{Experimental Setup}

\subsection{Artifacts and Ground Truth}

We reused the 30 artifacts and 150 ground-truth errors from the CCR experiment \citep{song2026ccr}: 10 code artifacts (C1--C10), 10 technical documents (D1--D10), and 10 presentation scripts (S1--S10), each containing exactly 5 injected errors across five types (FACT, CONS, CTXT, RCVR, MISS) at three severity levels.

\subsection{Execution}

All sessions used Claude Opus 4.6 via the \texttt{claude} CLI with the \texttt{-p} flag, which launches each invocation as an independent session with no context carryover.
This provides natural context separation: each CLI call is functionally equivalent to a fresh API session.
The CLI uses default API parameters: temperature = 1.0 (Claude API default), no top-$p$ override, no max token limit override (model default applies).
No custom sampling parameters were set.

For the main experiment, we ran CCR-1, D-CCR-2a, and D-CCR-2b with 3 independent runs per artifact ($30 \times 3 = 90$ sessions each), and D-CCR-2c with 1 run (30 sessions).
This totaled 300 experimental units (artifact $\times$ condition $\times$ run).
Each multi-turn unit required multiple CLI invocations (R1 review, question generation, author answer, R2 review), so the total number of CLI calls exceeded 300.
D-CCR-2c received only 1 run because pilot data showed its pattern was already clear, and D-CCR-3b (a three-round variant) was dropped after pilot results confirmed diminishing returns beyond round two.

The review prompt was in Korean, matching the artifact language.
This addresses the language confound noted in the original CCR experiment \citep{song2026ccr}, where fresh-session reviews defaulted to English.
In D-CCR, all conditions produced 94--96\% Korean-language output, effectively controlling for language effects.

\subsection{Prompt Design and Review Target Drift}

During pilot testing (10 artifacts $\times$ 5 conditions), we discovered a systematic failure mode in the D-CCR-2b prompt.
The initial prompt included ``also verify whether the author's answers are correct,'' which caused reviewers to shift from reviewing the artifact to critiquing the Q\&A conversation---generating findings about answer accuracy rather than artifact errors.
We call this \emph{Review Target Drift}.

In one representative pilot case, 3 of 4 new D-CCR-2b Round 2 findings were critiques of the author's answers (e.g., ``Q3's suggested fix has a type mismatch''), not artifact errors.
Since these critiques have no corresponding ground truth entry, they all count as false positives.

We revised the prompt (v2) to constrain the review target: ``The author's answers are for reference only.
Focus on finding new errors \emph{in the artifact itself}.
Do not critique the Q\&A exchange.'' This reduced the most egregious cases but did not eliminate the FP increase, suggesting a structural component beyond prompt wording.

The v1$\to$v2 prompt change constitutes an ablation on review target specification; we discuss its implications in Section~\ref{sec:discussion}.

\subsection{Matching Algorithm}

We match reviewer findings to ground-truth errors using a scoring function that combines: (1) line number proximity ($\pm$5 lines), (2) keyword overlap with Korean morphological normalization (suffix stripping for agglutinative particles), and (3) fuzzy substring matching.
A finding counts as a true positive if its score exceeds 2.0.

Korean suffix stripping was added after independent verification identified that keyword matching underperformed on Korean text due to agglutinative morphology.
The relative ordering of conditions was consistent across all thresholds tested (1.0--3.0).

The D-CCR CCR-1 baseline (F1 = 0.376) exceeds the original CCR experiment's fresh-session result (F1 = 0.286; \citealp{song2026ccr}).
This reflects the v2 matching algorithm (Korean suffix stripping + fuzzy substring matching), not a change in review quality---re-analyzing the original CCR data with v2 yields comparable values.
All conditions use the same matching algorithm, so relative comparisons are valid.

\subsection{Statistical Methods}

Precision is defined as TP / Findings, where Findings = TP + FP + Dup (see Section~\ref{sec:results}).
This means duplicate findings---those matching a ground-truth error already identified by another finding---count against precision.
This is conservative: it penalizes verbose reviewers who describe the same error multiple times with different wording.

We compute per-artifact F1 scores averaged across 3 runs (1 run for D-CCR-2c), yielding 30 paired observations per comparison.
We report paired $t$-tests with Cohen's $d$ for effect size, supplemented by Wilcoxon signed-rank tests for non-parametric confirmation.
We conduct four planned comparisons (H1, H3, H4, H5); applying Bonferroni correction ($\alpha = 0.05/4 = 0.0125$), all significant results remain significant (all $p < 0.001$ except H3, which is non-significant regardless of correction).

\section{Results}
\label{sec:results}

\subsection{Overall Performance}

Table~\ref{tab:main} shows the main results, with per-artifact scores averaged across runs.

\begin{table*}[t]
\centering
\caption{Main experiment results ($N = 30$ artifacts).
Findings = TP + FP + Dup, where Dup counts findings matching a ground-truth error already identified by another finding (redundant true positives).
Duplicate rates increase with rounds (15\% for CCR-1 vs.\ 30\% for D-CCR-2c).}
\label{tab:main}
\begin{tabular}{lcccccccc}
\toprule
\textbf{Condition} & \textbf{Findings} & \textbf{TP} & \textbf{FP} & \textbf{Dup} & \textbf{Precision} & \textbf{Recall} & \textbf{F1} & \textbf{F1 SD} \\
\midrule
\textbf{CCR-1} & 9.3 & 2.64 & 5.23 & 1.43 & \textbf{0.297} & 0.529 & \textbf{0.376} & 0.136 \\
D-CCR-2a & 15.4 & 2.96 & 9.17 & 3.27 & 0.197 & 0.591 & 0.293 & 0.102 \\
D-CCR-2b & 15.2 & 3.03 & 8.47 & 3.70 & 0.204 & 0.607 & 0.303 & 0.110 \\
D-CCR-2c & 18.4 & 3.10 & 9.70 & 5.60 & 0.168 & \textbf{0.620} & 0.263 & 0.091 \\
\bottomrule
\end{tabular}
\end{table*}

\begin{figure}[t]
\centering
\includegraphics[width=\columnwidth]{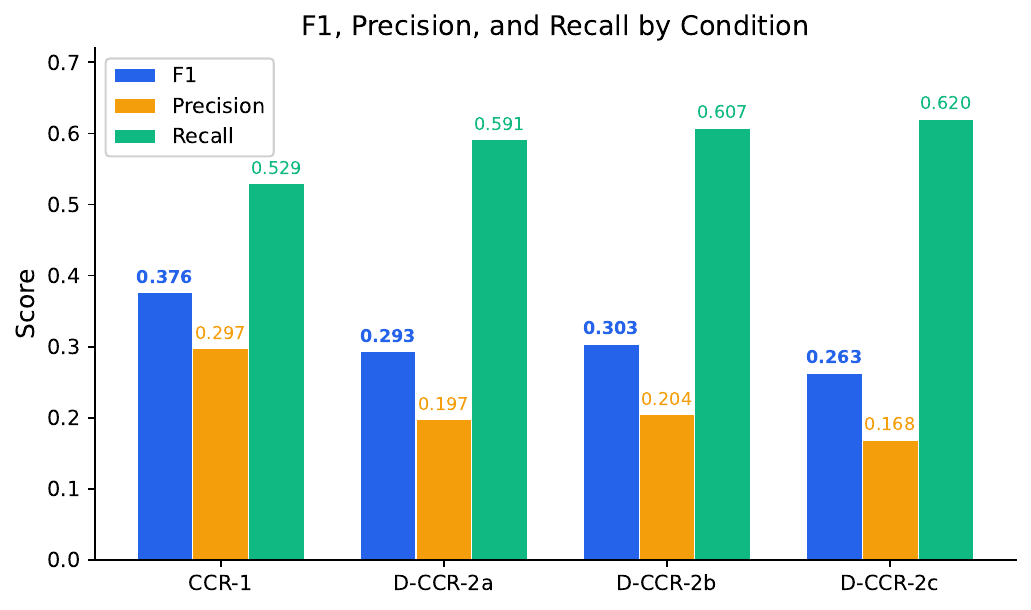}
\caption{F1, Precision, and Recall by condition. CCR-1 achieves the highest F1 and Precision. Multi-turn conditions increase Recall but collapse Precision, resulting in lower F1.}
\label{fig:metrics}
\end{figure}

CCR-1 leads on F1 and precision.
Multi-turn conditions produce 63--97\% more findings but with substantially lower precision.
D-CCR-2b finds slightly more true positives (3.03 vs.\ 2.64) but generates 62\% more false positives (8.47 vs.\ 5.23), yielding a significant net F1 decrease.
Figure~\ref{fig:metrics} visualizes the tradeoff: multi-turn conditions gain recall at devastating cost to precision and F1.

\subsection{Hypothesis Tests}

Table~\ref{tab:tests} reports statistical comparisons.

\begin{table*}[t]
\centering
\caption{Paired tests ($N = 30$). All significant results survive Bonferroni correction ($\alpha = 0.0125$).}
\label{tab:tests}
\resizebox{\textwidth}{!}{%
\begin{tabular}{lcccccccl}
\toprule
\textbf{Comparison} & $\Delta$\textbf{F1} & \textbf{95\% CI} & \textbf{Cohen's} $d$ & $t$ & $p$ & \textbf{Wilcoxon} $z$ & $p$ & \textbf{Dir.} \\
\midrule
H1: 2b vs.\ CCR-1 & $-0.073$ & [$-0.105$, $-0.040$] & $-0.59$ & $-4.57$ & ${<}0.001$ & $-3.63$ & ${<}0.001$ & \textbf{Rev.} \\
H3: 2b vs.\ 2a & $+0.010$ & [$-0.014$, $+0.034$] & $+0.09$ & $0.85$ & $0.20$ & $-0.89$ & $0.20$ & ns \\
H4: 2b vs.\ 2c & $+0.040$ & [$+0.018$, $+0.062$] & $+0.39$ & $3.70$ & ${<}0.001$ & $-3.16$ & ${<}0.01$ & Conf. \\
H5: 2c vs.\ CCR-1 & $-0.113$ & [$-0.149$, $-0.076$] & $-0.97$ & $-6.34$ & ${<}0.001$ & $-4.23$ & ${<}0.001$ & \textbf{Rev.} \\
\bottomrule
\end{tabular}%
}
\end{table*}

\textbf{H1 is reversed.} D-CCR-2b significantly underperforms CCR-1 ($p < 0.001$, $d = -0.59$). Multi-turn review is worse than single-pass.

\textbf{H3 is null.} Including author answers (2b) vs.\ excluding them (2a) makes no significant difference ($p \approx 0.20$, $d = 0.09$). The Q\&A exchange provides negligible benefit.

\textbf{H4 is confirmed.} Continuity (2b) outperforms independence (2c) ($p < 0.001$, $d = 0.39$).
Seeing prior questions and answers is better than starting fresh, but both are worse than reviewing only once.

\textbf{H5 is reversed.} Independent repetition (2c) is significantly worse than single-pass ($p < 0.001$, $d = -0.97$).
Two independent looks are worse than one.
This is the largest effect size in the study.

\subsection{The Recall--Precision Tradeoff}

All multi-turn conditions achieve higher recall than CCR-1: D-CCR-2c reaches 0.620 (+17\% over CCR-1's 0.529).
Additional rounds do find more true errors.
But the recall gain comes at severe cost to precision: CCR-1's 0.297 drops to 0.168 for D-CCR-2c---a 43\% decrease.

For every additional true positive discovered in round two, multi-turn review generates approximately 4--5 additional false positives. The signal-to-noise ratio worsens with each round.

\subsection{Marginal Discovery Rate}

We define the Marginal Discovery Rate (MDR) as the average across all 30 artifacts of the per-artifact ratio: $\text{MDR} = \mathbb{E}[\text{R2\_new\_TP}_i / \text{R1\_TP}_i]$, where $i$ indexes artifacts.
For artifacts where R1 found zero true positives ($\text{R1\_TP} = 0$), the ratio is defined as 0.
This convention is conservative: excluding zero-discovery artifacts would inflate MDR.

\begin{table}[t]
\centering
\caption{Round 2 discovery metrics.
R1 TP values vary slightly across conditions (2.61--2.67) due to probabilistic model behavior.
N(R2 new${>}$0) shows how many artifacts had any new true positive in round 2.}
\label{tab:mdr}
\small
\begin{tabular}{lccccc}
\toprule
\textbf{Cond.} & \textbf{R1 TP} & \textbf{R2 New} & \textbf{MDR} & \textbf{N(${>}$0)} & \textbf{R2 FP$\uparrow$} \\
\midrule
2a & 2.61 & 0.34 & 0.153 & 20/30 & +3.94 \\
2b & 2.67 & 0.37 & 0.188 & 15/30 & +3.24 \\
2c & 2.67 & 0.43 & 0.286 & 9/30 & +4.47 \\
\bottomrule
\end{tabular}
\end{table}

D-CCR-2b discovers 0.37 new TPs in round 2 (MDR = 0.188) but adds 3.24 new FPs.
The ``exchange rate'' is roughly 1 new TP for every 9 new FPs.
Round 2 contributed at least one new TP in only 15 of 30 artifacts---meaning half the time, the additional round produced nothing but noise.

D-CCR-2c has the highest MDR (0.286) because its completely independent round 2 rediscovers many round-1 errors plus some new ones.
But it also generates the most new FPs (4.47), making it the worst performer overall despite the highest MDR.

\subsection{Performance by Artifact Type}

\begin{table}[t]
\centering
\caption{F1 by artifact category.}
\label{tab:category}
\small
\begin{tabular}{lcccc}
\toprule
\textbf{Category} & \textbf{CCR-1} & \textbf{2a} & \textbf{2b} & \textbf{2c} \\
\midrule
Code & 0.354 & 0.308 & 0.284 & 0.255 \\
Document & 0.446 & 0.334 & 0.361 & 0.318 \\
Script & 0.328 & 0.238 & 0.265 & 0.217 \\
\bottomrule
\end{tabular}
\end{table}

CCR-1 leads across all three categories.
The gap is smallest for documents and largest for scripts.
Document artifacts show the highest absolute F1 values, consistent with the original CCR finding that documents contain more discoverable errors.

\subsection{Per-Artifact Analysis}

CCR-1 achieved the best F1 in 25 of 30 artifacts.
The five exceptions (C1, C4, D1, S4, S10) involved artifacts where CCR-1's round-1 recall was unusually low, giving multi-turn conditions room for marginal TP gains.
No artifact showed multi-turn review consistently outperforming across multiple conditions.

\subsection{Reproducibility}

Intra-artifact F1 standard deviation across 3 runs was 0.094 for CCR-1, 0.065 for D-CCR-2a, and 0.064 for D-CCR-2b (D-CCR-2c had 1 run).
The lower variance in multi-turn conditions reflects a floor effect compressing the distribution.
All 300 sessions used the same model snapshot (Claude Opus 4.6), prompts, and matching algorithm (v2).
Session-level outputs and analysis scripts are available for replication.

\section{Discussion}
\label{sec:discussion}

\subsection{Why More Rounds Fail: False Positive Pressure}

The primary mechanism is \emph{false positive pressure}.
After round one, the ``easy'' errors have been found.
Round two asks the reviewer to find \emph{new} errors, but the artifact may have few or no remaining discoverable errors.
Under pressure to produce findings, the reviewer generates speculative, low-confidence findings that do not correspond to real errors.

\begin{figure}[t]
\centering
\includegraphics[width=\columnwidth]{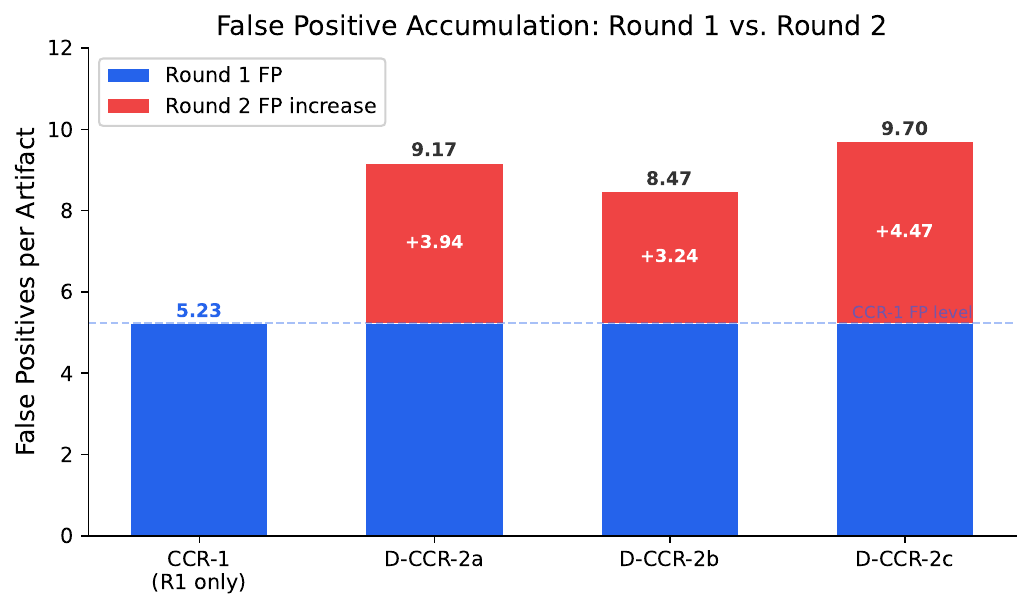}
\caption{False positive accumulation across conditions. The dashed line marks CCR-1's FP level. All multi-turn conditions add substantial Round 2 FPs (red), with D-CCR-2c generating the most noise.}
\label{fig:fp}
\end{figure}

The numbers are clear: CCR-1 produces 9.3 findings per artifact; D-CCR-2b produces 15.2---a 63\% increase.
True positives increase only from 2.64 to 3.03 (15\%).
The rest is false positives: 5.23 to 8.47.
Figure~\ref{fig:fp} shows the FP accumulation pattern: Round 2 adds 3--4.5 new false positives per artifact across all multi-turn conditions.

 For artifact C3 in D-CCR-2b Run 1, the reviewer found 3 true errors in round 1 but generated 7 new findings in round 2, of which only 1 was genuine.
The remaining 6 included speculative performance concerns, style suggestions, and hypothetical edge cases.

This mirrors \citeauthor{song2026ccr}'s (\citeyear{song2026ccr}) SR2 finding: a second review pass increased findings but decreased precision.
The mechanism is the same whether the second pass happens in the same session or a separated one.
Reviewing again creates pressure to find something new, and when nothing new exists, the model invents it.

\subsection{Review Target Drift}

The pilot revealed a more specific failure mode in D-CCR-2b.
When the reviewer receives the Q\&A exchange, it sometimes shifts from reviewing the artifact to reviewing the conversation---flagging, for example, ``the author's answer to Q3 contains a type mismatch in the suggested fix.'' This is a valid observation about the answer, but not an error in the original artifact.

Our v2 prompt explicitly constrained the review target.
The improvement was modest (pilot v1 FP = 11.4, v2 FP = 10.5 for D-CCR-2b), suggesting a structural component: when conversation content is present, the model's attention distributes across it regardless of instructions.

For practitioners implementing multi-turn LLM review: either carefully distinguish artifact findings from conversation findings in the output, or restrict the reviewer to the artifact plus a structured summary of prior findings rather than the raw conversation.

\subsection{The Round Optimization Principle}

The central theoretical contribution is that the optimal number of CCR review rounds is exactly one---not as a practical heuristic, but as an empirical finding with a clear mechanism.

The data does \emph{not} support a simple ``less information is better'' narrative.
Within multi-turn conditions, more information helps: D-CCR-2b (artifact + Q\&A) outperforms D-CCR-2a (artifact + questions only), which outperforms D-CCR-2c (artifact only, fresh):

\begin{enumerate}
\item \textbf{D-CCR-2b} (Q\&A exchange): F1 = 0.303
\item \textbf{D-CCR-2a} (questions only): F1 = 0.293
\item \textbf{D-CCR-2c} (no prior info): F1 = 0.263
\end{enumerate}

Given that a second round happens, more context is better.
The problem is that \emph{any} second round---regardless of accompanying information---degrades performance compared to stopping after one.
The mechanism is false positive pressure: reviewing again creates demand for new findings that the artifact cannot supply.

This distinguishes D-CCR's finding from CCR's original insight.
CCR showed that \emph{removing production context} improves review (information restriction).
D-CCR shows that \emph{adding review rounds} degrades review (round count).
These are complementary but distinct principles.
The optimal strategy: separate the context, review once, and if more compute is available, run independent parallel reviews rather than sequential ones.

This aligns with multi-agent debate findings.
\citet{zhang2025mad} found that MAD fails without model heterogeneity---the same model with different prompts is insufficient.
\citet{wu2025debate} showed that interaction creates majority pressure suppressing independent correction.
D-CCR's round-2 degradation likely reflects a similar dynamic: the reviewer's own prior output constrains what the second review can contribute.

\subsection{Practical Implication: Ensemble Over Iteration}

If the goal is to spend more compute on verification, \emph{parallel independent reviews} (ensemble) are preferable to \emph{sequential iterative reviews} (D-CCR).
A majority-vote ensemble of 3 independent CCR-1 reviews (accepting a finding only if 2 of 3 runs agree) achieved F1 = 0.393, outperforming D-CCR-2b's F1 = 0.303 ($p < 0.001$).
The ensemble approach avoids both false positive pressure and Review Target Drift because each reviewer works independently with no knowledge of other reviews.

\subsection{Implications for the CCR Research Program}

D-CCR was the second planned study in a four-paper research program \citep[Section 8]{song2026ccr}.
The hypothesis that multi-turn review improves on static CCR was falsified, but the negative result establishes a boundary condition: context separation is necessary and sufficient; iterative interaction within separated contexts does not add value.

This redirects the research program.
Sequential rounds with a single reviewer do not help, but the next study---Hierarchical Cross-Context Aggregation (HCCA)---takes a different approach: structurally dividing the review task across specialized agents rather than repeating the same review.
D-CCR's failure is a failure of \emph{iteration with a single role}, not of \emph{multi-agent architectures} where agents contribute structurally distinct analyses.
Improving what happens \emph{within} the single review round---better prompting, structured checklists, or domain-specific protocols---also remains productive while maintaining the single-pass architecture.

\section{Limitations}

\textbf{Single model.} All experiments used Claude Opus 4.6. Models with different calibration or verbosity tendencies may show different recall-precision tradeoffs in multi-round settings.

\textbf{Injected errors.} Errors were deliberately injected, not naturally occurring.
Natural errors may be more varied and harder to detect, potentially creating more room for multi-round discovery, though our injected errors span five types and three severity levels.

\textbf{Matching algorithm.} Our heuristic matching (line proximity + keyword overlap with Korean suffix stripping) produces consistent relative ordering across thresholds 1.0--3.0, though absolute F1 values are threshold-dependent.
Manual verification of 20 borderline cases confirmed all were genuine false positives, but the matching has known limitations with paraphrased error descriptions.

\textbf{S5 session crashes.} Four sessions for artifact S5 (two D-CCR-2b runs, two D-CCR-2a runs) produced incomplete outputs due to CLI crashes.
These were excluded from run averaging.
The $N = 30$ artifact-level analysis is unaffected as at least one valid run exists for each condition.

\textbf{D-CCR-2c single run.} D-CCR-2c used only 1 run (vs.\ 3 for other conditions), reducing precision of its estimates.
However, the effect size ($d = -0.97$ vs.\ CCR-1) is large enough that the conclusion is robust.

\textbf{D-CCR-2c pooling.} D-CCR-2c pools findings from both R1 and R2, giving it twice the opportunity to generate findings compared to CCR-1's single round.
This structural asymmetry means D-CCR-2c's higher finding count (18.4 vs.\ 9.3) partly reflects more attempts rather than better detection.
The comparison remains valid for its intended purpose---testing whether ``two independent looks'' beat one---but should be interpreted with this asymmetry in mind.

\textbf{No human baseline.} We did not compare against human reviewers, which would provide an upper bound on error detection performance.

\section{Ethics and Broader Impact}

This study uses only synthetic artifacts with deliberately injected errors; no human subjects or personal data were involved.
Our results apply specifically to LLM-as-reviewer settings and should not be generalized to human review processes, where multi-round review has established benefits.
The false positive pressure phenomenon may vary across models and domains; practitioners should validate these findings in their context before adopting single-pass review protocols.

\section{Conclusion}

Multi-turn review does not improve Cross-Context Review.
Single-pass CCR (F1 = 0.376) significantly outperformed all multi-turn variants (F1 = 0.263--0.303).
Additional rounds increase recall modestly but generate false positives at a much higher rate, destroying precision.
The optimal number of review rounds is one.

False positive pressure and Review Target Drift explain why ``more review should help'' fails in practice.
These findings complement CCR's original insight: while CCR showed that \emph{removing production context} improves review, D-CCR shows that \emph{adding review rounds} degrades it.
Within multi-turn conditions, more information helps (D-CCR-2b $>$ D-CCR-2a $>$ D-CCR-2c)---the problem is round count, not information amount.

For practitioners with compute budget for additional verification: spend it on independent parallel reviews (ensemble), not sequential iterative ones.
Separate the context, review once, and trust the first answer.

\bibliography{references}

@inproceedings{chen2025magicore,
  title={{MAgICoRe}: Multi-Agent, Iterative, Coarse-to-Fine Refinement for Reasoning},
  author={Chen, Justin Chih-Yao and Prasad, Archiki and Saha, Swarnadeep and Stengel-Eskin, Elias and Bansal, Mohit},
  booktitle={Proceedings of the 2025 Conference on Empirical Methods in Natural Language Processing (EMNLP 2025)},
  year={2025}
}

@inproceedings{du2025thinkthrice,
  title={Think Thrice Before You Act: Progressive Thought Refinement in Large Language Models},
  author={Du, Chengyu and Han, Jinyi and Ying, Yueqi and Chen, Aili and He, Qian and Zhao, Haotian and Xia, Shunian and Guo, Haoyang and Liang, Jie and Chen, Zhi and Li, Lei and Xiao, Yanghua},
  booktitle={Proceedings of the 13th International Conference on Learning Representations (ICLR 2025)},
  year={2025}
}

@inproceedings{gou2024critic,
  title={{CRITIC}: Large Language Models Can Self-Correct with Tool-Interactive Critiquing},
  author={Gou, Zhibin and Shao, Zhihong and Gong, Yeyun and Shen, Yelong and Yang, Yujiu and Duan, Nan and Chen, Weizhu},
  booktitle={Proceedings of the 12th International Conference on Learning Representations (ICLR 2024)},
  year={2024}
}

@inproceedings{guo2025temporal,
  title={Temporal Consistency for {LLM} Reasoning Process Error Identification},
  author={Guo, Jiacheng and Wu, Yiming and Qiu, Jiaqi and Huang, Keyu and Juan, Xin and Yang, Liangliang and Wang, Maozheng},
  booktitle={Findings of the Association for Computational Linguistics: EMNLP 2025},
  pages={22114--22129},
  year={2025}
}

@inproceedings{huang2024llms,
  title={Large Language Models Cannot Self-Correct Reasoning Yet},
  author={Huang, Jie and Chen, Xinyun and Mishra, Swaroop and Zheng, Huaixiu Steven and Yu, Adams Wei and Song, Xinying and Zhou, Denny},
  booktitle={Proceedings of the 12th International Conference on Learning Representations (ICLR 2024)},
  year={2024}
}

@article{kamoi2024when,
  title={When Can {LLM}s Actually Correct Their Own Mistakes? {A} Critical Survey of Self-Correction of {LLM}s},
  author={Kamoi, Ryo and Zhang, Yusen and Zhang, Nan and Han, Jiawei and Zhang, Rui},
  journal={Transactions of the Association for Computational Linguistics},
  volume={12},
  pages={1417--1440},
  year={2024}
}

@inproceedings{kim2025interviewer,
  title={{LLM}-as-an-Interviewer: Beyond Static Testing Through Dynamic {LLM} Evaluation},
  author={Kim, Eunsu and Suk, Juyoung and Kim, Seungone and Muennighoff, Niklas and Kim, Dongkwan and Oh, Alice},
  booktitle={Findings of the Association for Computational Linguistics: ACL 2025},
  year={2025}
}

@inproceedings{kumar2024score,
  title={Training Language Models to Self-Correct via Reinforcement Learning},
  author={Kumar, Aviral and Zhuang, Vincent and Agarwal, Rishabh and Su, Yi and Co-Reyes, John D. and Singh, Avi and Baumli, Kate and Iqbal, Shariq and Bishop, Colton and Roelofs, Rebecca and Zhang, Lei M. and McKinney, Kay and Shrivastava, Disha and Paduraru, Cosmin and Tucker, George and Precup, Doina and Behbahani, Feryal and Faust, Aleksandra},
  booktitle={Proceedings of the 13th International Conference on Learning Representations (ICLR 2025, Oral)},
  year={2025}
}

@article{shi2025ssr,
  title={{SSR}: Socratic Self-Refine for Large Language Model Reasoning},
  author={Shi, Hao and others},
  journal={arXiv preprint arXiv:2511.10621},
  year={2025}
}

@inproceedings{smit2024mad,
  title={Should We Be Going {MAD}? {A} Look at Multi-Agent Debate Strategies for {LLM}s},
  author={Smit, Andries and others},
  booktitle={Proceedings of the 41st International Conference on Machine Learning (ICML 2024)},
  pages={45883--45905},
  year={2024}
}

@article{song2026ccr,
  title={Cross-Context Review: Improving {LLM} Output Quality by Separating Production and Review Sessions},
  author={Song, Tae-Eun},
  journal={arXiv preprint arXiv:2603.12123},
  year={2026}
}

@article{tsui2025selfcorrection,
  title={Self-Correction Bench: Uncovering and Addressing the Self-Correction Blind Spot in Large Language Models},
  author={Tsui, Kenneth},
  journal={arXiv preprint arXiv:2507.02778},
  year={2025}
}

@article{wu2025debate,
  title={Can {LLM} Agents Really Debate? {A} Controlled Study of Multi-Agent Debate in Logical Reasoning},
  author={Wu, Hao and Li, Zhiwei and Li, Ling},
  journal={arXiv preprint arXiv:2511.07784},
  year={2025}
}

@inproceedings{zhang2025dark,
  title={Understanding the Dark Side of {LLM}s' Intrinsic Self-Correction},
  author={Zhang, Qingjie and Wang, Dingjie and Qian, Hao and Li, Yixuan and Zhang, Tian and Huang, Minlie and Xu, Kai and Li, Hao and Liu, Yang and Qiu, Haipeng},
  booktitle={Proceedings of the 63rd Annual Meeting of the Association for Computational Linguistics (ACL 2025)},
  year={2025}
}

@article{zhang2025mad,
  title={If Multi-Agent Debate is the Answer, What is the Question?},
  author={Zhang, Hangfan and Cui, Zhimeng and Chen, Jinghan and Wang, Xiang and Zhang, Qiang and Wang, Ziyue and Wu, Ding and Hu, Shudong},
  journal={arXiv preprint arXiv:2502.08788v2},
  year={2025}
}

\appendix

\section{Pilot Experiment and Prompt Evolution}
\label{app:pilot}

\subsection{Pilot Design and H2 Drop Decision}

Hypothesis numbering follows the original pre-registration.
H2 tested whether three rounds outperform two (D-CCR-3b vs.\ D-CCR-2b).
Pilot results showed D-CCR-3b (F1 = 0.216) was substantially worse than D-CCR-2b (F1 = 0.270), confirming diminishing returns beyond round two.
D-CCR-3b was dropped from the main experiment, and H2 is not reported.

We ran a pilot experiment with 10 artifacts $\times$ 5 conditions (including D-CCR-3b) $\times$ 1 run = 50 sessions. Pilot results:

\begin{table}[h]
\centering
\small
\begin{tabular}{lccc}
\toprule
\textbf{Condition} & \textbf{F1} & \textbf{Precision} & \textbf{Recall} \\
\midrule
CCR-1 & 0.366 & 0.307 & 0.480 \\
D-CCR-2a & 0.258 & 0.173 & 0.520 \\
D-CCR-2b & 0.270 & 0.184 & 0.520 \\
D-CCR-2c & 0.241 & 0.154 & 0.560 \\
D-CCR-3b & 0.216 & 0.137 & 0.540 \\
\bottomrule
\end{tabular}
\end{table}

D-CCR-3b confirmed diminishing returns: three rounds performed worse than two. This condition was dropped from the main experiment.

\subsection{Prompt v1 $\to$ v2}

The pilot D-CCR-2b prompt (v1) included: ``also verify whether the author's answers are correct.'' This caused Review Target Drift.
The v2 prompt replaced this with: ``The author's answers are for reference only.
Focus on finding new errors in the artifact itself.''

FP reduction from v1 to v2 was modest (11.4 $\to$ 10.5 for D-CCR-2b), confirming that the FP problem is structural, not merely a prompt issue.

\subsection{Matching Algorithm Evolution}

The pilot initially used exact keyword matching, which underperformed on Korean text.
Independent verification identified that agglutinative suffixes caused match failures.
We added suffix stripping and fuzzy substring matching (v2 algorithm), then re-analyzed all data.
The H1 reversal (CCR-1 $>$ D-CCR-2b) was robust across both matching versions and all thresholds tested (1.0--3.0).

\section{Threshold Robustness Analysis}

The relative ordering CCR-1 $>$ D-CCR-2b holds at every matching threshold from 1.0 to 3.0, as independently verified.
At threshold 1.0 (most permissive), CCR-1 recall exceeds D-CCR-2b recall, eliminating the possibility that our results depend on a threshold that favors CCR-1's matching patterns.

\section{Ensemble Analysis}

A majority-vote ensemble (count a finding as TP if $\geq$2 of 3 runs agree) yields:

\begin{table}[h]
\centering
\small
\begin{tabular}{lc}
\toprule
\textbf{Condition} & \textbf{Ensemble F1} \\
\midrule
CCR-1 (Majority 2/3) & 0.393 \\
D-CCR-2b (Majority 2/3) & 0.311 \\
\bottomrule
\end{tabular}
\end{table}

The ensemble CCR-1 result (0.393) is the highest F1 achieved in this study, confirming that parallel independent reviews outperform sequential iterative ones.

\end{document}